\documentclass[letterpaper]{article}
\usepackage[preprint]{aaai2027}
\usepackage[hyphens]{url}
\usepackage{graphicx}
\usepackage{natbib}
\usepackage{caption}
\usepackage{algorithm}
\usepackage{algorithmic}

\usepackage{newfloat}
\usepackage{listings}
\DeclareCaptionStyle{ruled}{labelfont=normalfont,labelsep=colon,strut=off}
\floatstyle{ruled}
\newfloat{listing}{tb}{lst}{}
\floatname{listing}{Listing}

\usepackage{booktabs}

\usepackage{amsfonts}
\usepackage{amsmath}
\usepackage{booktabs}
\usepackage{multirow}

\title{Beyond Clean Data: Robust Test-Time Adaptation on Noisy Time-Series Streams}

\author{
    Kaixin Wang\textsuperscript{\rm 1},
    Jinping Wang\textsuperscript{\rm 1},
    Zhiqiang Gao\textsuperscript{\rm 1}
}

\affiliations{
    \textsuperscript{\rm 1}Wenzhou-Kean University\\
    1306318@wku.edu.cn, 1306325@wku.edu.cn, zgao@wku.edu.cn
}

\begin{document}

\maketitle

\begin{abstract}
Test-time adaptation for time-series classification aims to address domain shifts at inference time using only streaming unlabeled target data. Although many time-series TTA methods have shown promise, some remain unstable under contaminated streaming conditions. However, existing time-series TTA protocols do not explicitly evaluate robustness under such conditions. To illustrate this limitation, we report the performance of existing methods under five contaminated-stream protocols involving near- and far-distribution OOD samples, adversarial samples, noise samples, and perturbation samples during adaptation, with performance measured on benign target samples. To improve robustness under these conditions, we propose a quality-weighted Global Gravitational Potential Loss (GGPot) that does not use class-valued pseudo labels to define sample relations. It models batch samples as interacting particles and optimizes a quality-weighted global potential through attraction and collapse-preventing repulsion. Comparisons with representative TTA approaches on real-world time-series datasets under five contaminated-stream settings show that ACCUP with GGPot improves the average performance over the original ACCUP and achieves the highest average performance among the evaluated methods, demonstrating the effectiveness of GGPot for robust time-series TTA.
\end{abstract}

\section{Introduction}

Domain shifts between training and test data severely degrade the generalization capability of time series models in real-world scenarios. To address domain shifts, unsupervised domain adaptation (UDA) methods \cite{58_wilson2020survey} have been developed to align feature representations in the embedding space by leveraging unlabeled target data \cite{4_cai2021time, 15_he2023domain, 19_jin2022domain, 41_ragab2022self, 42_ragab2023adatime, 59_wilson2020multi, 60_wilson2023calda}. Nevertheless, the reliance on source domain data availability renders these approaches impractical in privacy-sensitive scenarios. To overcome this limitation, source-free domain adaptation (SFDA) provides a promising alternative that operates exclusively on the target domain data \cite{24_kim2021domain, 43_ragab2023source, 45_ren2023single, 62_xiao2023temporal}. However, SFDA struggles to perform immediate inference with streaming data in online time series applications. For example, deploying a pre-trained human activity recognition model to a new person requires collecting a complete target dataset, which is incompatible with dynamic streaming environments.

A more practical approach lies in test-time adaptation (TTA), which enables online adaptation with streaming data \cite{29_liang2025comprehensive}. By avoiding the high training costs associated with SFDA, TTA ensures continuous adaptation, making it suitable for scenarios requiring immediate inference. Recent advancements in TTA applications for time series classification tasks have shown promising progress. Existing methods improve adaptation from several perspectives. Some methods rely on confidence-oriented objectives to reduce prediction uncertainty under distribution shifts \cite{12_gong2022note, 38_niu2022efficient, 39_niu2023towards, 46_schneider2020improving, 54_wang2020tent, 55_wang2022continual, 65_yuan2023robust}. Some others instead leverage prediction-based self-training signals to construct adaptation supervision from the model's own outputs \cite{25_lee2013pseudo, 30_liang2020we, 48_su2022revisiting}. In addition, several methods further exploit structural cues in the target domain, such as support sets, nearest neighbors, prototypes, or clustering, to improve the reliability of adaptation signals \cite{17_iwasawa2021test, 18_jang2022test, 56_wang2023feature, 005_gong2025augmented}. Nevertheless, these methods are starting to show diminishing returns, and their performance in certain scenarios is unstable. A key reason is that TTA is highly sensitive to noisy and shifted target-domain data, which can easily corrupt adaptation signals and thus exacerbate instability.

\begin{figure}[t]
    \centering
    \includegraphics[width=\columnwidth]{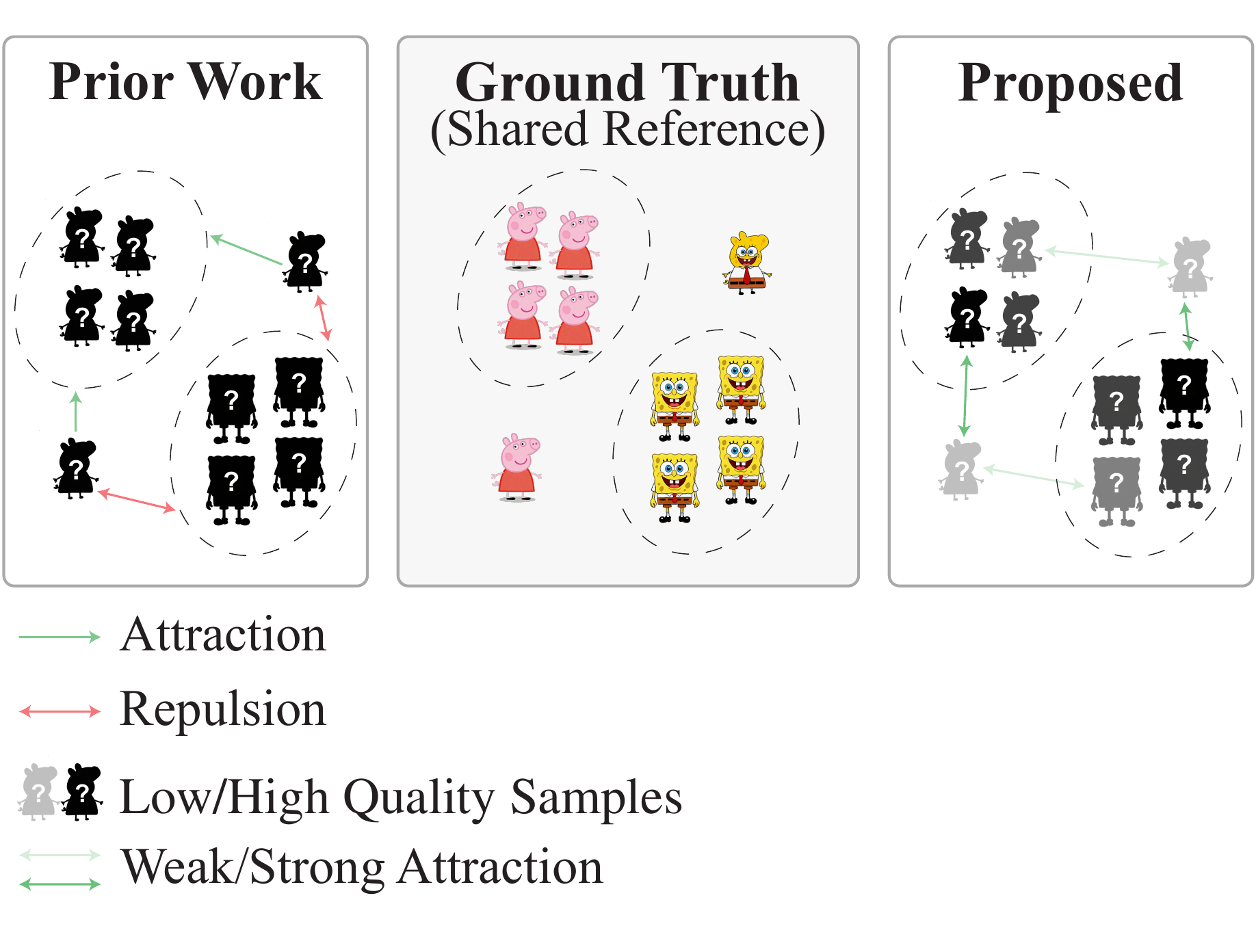}
    \caption{Left: pseudo-label-based adaptation can accumulate and propagate errors through incorrect pull-push relations; Middle: ground-truth semantic structure; Right: our method alleviates this issue with a quality-aware global potential field, without relying on pseudo labels.}
    \label{fig:error_propagation}
\end{figure}

In particular, many methods that explicitly optimize the structure of the target-domain feature space rely on pseudo labels to define supervision over target samples. This instability becomes particularly evident in such pseudo-label-based supervision, as shown in Fig.~\ref{fig:error_propagation}. Once some target samples are assigned incorrect pseudo labels, the induced pull-push relations no longer reflect the true semantic structure. As a result, samples from the same class may be pushed apart, while samples from different classes may be pulled together. Rather than remaining local, these incorrect relations can be repeatedly reinforced during online adaptation, leading to accumulated errors that propagate across the target domain.

Time series data often exhibit variations in magnitude and unavoidable noise \cite{014_duan2023effect}. Magnitude variability refers to amplitude changes in time series data across varying operating environments, sensor calibrations, inter-individual conditions, etc. A typical case in human activity recognition: although the semantic activity is unchanged, such as walking, the observed accelerometer magnitudes may vary due to different walking speeds. Beyond magnitude variability, real-world signals often exhibit noise and non-stationarity under environmental perturbations, which possibly distort the true signal dynamics and make pattern extraction more difficult \cite{015han2024time, 016_qin2024interpretable}.
Standard benchmark performance alone may not fully characterize the reliability of TTA in such environments. This raises an important question:
\textit{Can TTA remain stable under contaminated streaming conditions?}

To better evaluate the robustness of different methods, we consider five contaminated-stream protocols: Attack, Near OOD, Far OOD, Noise, and Perturb.
The Perturb. protocol further covers ten representative time-series perturbations at five severity levels.
These protocols allow us to examine whether existing methods remain stable across various forms of stream contamination.

Under the proposed protocols, existing TTA methods exhibit markedly different levels of stability across contaminated streams. This result supports our central insight that robustness under contaminated streaming conditions provides an important complementary perspective to average performance under standard settings. In particular, methods that leverage pseudo-label-driven contrastive learning may obtain noticeable improvements in standard benchmarks, but such improvements can be harder to sustain in more realistic and error-prone settings. Motivated by this observation, \textbf{we propose \underline{G}lobal \underline{G}ravitational \underline{Pot}ential Loss (GGPot)}, a pseudo-label-free, quality-weighted, gravitationally inspired objective that treats test-time batch samples as interacting particles and minimizes their global potential rather than relying on contrastive objectives. Experiments on two time-series benchmarks show that our method achieves the highest average and worst-case performance among the evaluated methods under the five contaminated-stream protocols.

The main contributions of this paper are as follows:
\begin{itemize}
\item We revisit time-series TTA from a robustness-oriented perspective and complement standard cross-domain evaluation with contaminated-stream protocols that assess stability under various adaptation interference.
\item To the best of our knowledge, this work is among the first to systematically evaluate time-series TTA under contaminated adaptation streams.
\item We propose \textbf{GGPot}, a pseudo-label-free, quality-weighted objective that reduces reliance on potentially erroneous pseudo-label supervision by optimizing the system-level potential over batch samples, achieving the highest average and worst-case performance among the evaluated methods under the five contaminated-stream protocols.
\end{itemize}

\section{Related Work}
\paragraph{Time Series Domain Adaptation}
Unsupervised Domain Adaptation (UDA) has been widely studied to mitigate performance degradation caused by domain shifts requiring labeled target data. Current approaches for time series UDA mainly fall into two main categories, discrepancy-based alignment and adversarial learning \cite{42_ragab2023adatime}. The former reduces statistical divergence between domains \cite{4_cai2021time,15_he2023domain,34_liu2021adversarial}, and the latter learns domain-invariant representations via adversarial objectives \cite{19_jin2022domain, 41_ragab2022self, 59_wilson2020multi,60_wilson2023calda}. Despite notable progress, prior time-series adaptation methods are mainly developed for standard domain-shift settings and are less concerned with robustness under noisy streaming target data.

\paragraph{Test-Time Adaptation}
Test-time Adaptation (TTA) addresses domain shift by adapting a pre-trained model using only unlabeled test data during inference. One line of work is auxiliary self-supervised tasks that are applied during both training and testing phases \cite{35_liu2021ttt++,51_sun2020test}, though such approaches may be difficult to deploy in practice. A more common paradigm focuses on adapting models solely at test time using online unlabeled data streams \cite{3_brahma2023probabilistic,6_chen2022contrastive,14_goyal2022test,40_press2023rdumb,46_schneider2020improving,54_wang2020tent}.
Other methods incorporate techniques such as feature alignment or prototype-based learning to further improve adaptation performance \cite{17_iwasawa2021test,18_jang2022test,48_su2022revisiting,56_wang2023feature}.

\paragraph{Deep Metric Learning Loss}
Deep Metric Learning (DML) aims to learn a mapping from original samples to an embedding space, where samples from the same class are close to each other while samples from different classes are far apart. One of the main types of DML loss is ranking/pair-based DML loss, which often leverages the information derived from distance relationships across samples. Classical contrastive loss \cite{pfml_8_chopra2005learning} and triplet loss \cite{pfml_40_schroff2015facenet, pfml_52_weinberger2005distance, pfml_53_wu2017sampling} are typical representative instances. N-pair loss \cite{pfml_42_sohn2016improved} and lifted structured embedding \cite{pfml_43_oh2016deep} extend this idea by incorporating multiple negative samples into the training objective. Pair-based methods often rely on informative pair/tuple mining, usually based on sample difficulty, such as hard or semi-hard negatives and easy positives. GGPot naturally emphasizes local relations through distance decay, while also optimizing all sample interactions within a batch without tuple mining.

Potential fields have long been used to model interactions among objects \cite{pfml_20_khatib1985real} and have recently been extended to deep metric learning \cite{012_bhatnagar2025potential}. Unlike prior supervised potential-field formulations, our method is unsupervised and does not rely on class-dependent attraction and repulsion. Instead, GGPot models sample interactions through mutual attraction, analogous to gravity.

\section{Preliminary}
This paper considers a source-free streaming test-time adaptation setting for time-series classification. Let $\mathcal{D}_S=\{(X_S^i,y_S^i)\}_{i=1}^{n_S}$ denote the labeled source-domain dataset, and let $\mathcal{D}_T=\{X_T^i\}_{i=1}^{n_T}$ denote the unlabeled target-domain dataset. Each sample $X_S^i$ or $X_T^i$ denotes a univariate or multivariate time-series sequence of length $L$. $y_S^i$ denotes the label of the source sample $X_S^i$, and $n_S$ and $n_T$ are the sample sizes of the source and target domains. The source and target domains share the same label space but differ in data distributions. A source-pretrained predictor $g_{\theta_0}=h_{\phi_0}\circ f_{\psi_0}$ is first learned with standard empirical risk minimization on $\mathcal{D}_S$, where $f_{\psi_0}$ and $h_{\phi_0}$ denote the feature encoder and classifier, respectively.

In the streaming TTA setting, the target-domain data are not accessed all at once. Instead, $\mathcal{D}_T$ arrives sequentially as a stream of unlabeled batches $\{\mathcal{B}_b\}_{b=1}^{{N_b}},\quad \mathcal{B}_b=\{X_{T,b}^j\}_{j=1}^{{m_b}},$ where $N_b$ is the number of target batches, $m_b$ is the size of the $b$-th batch, and $\sum_{b=1}^{N_b}m_b=n_T$. For an input sample $X_{T,b}^j$, the model prediction is written as $p_b^j=g_{\theta}(X_{T,b}^j)\in\mathbb{R}^C,$ where $C$ is the number of classes.

In the standard cross-domain evaluation, among all domains $\{\mathcal{D}^{(k)}\}_{k=1}^M$ in a benchmark, a test-time adaptation scenario is an ordered pair $(s, t)$ defined by selecting one domain $\mathcal{D}^{(s)}$ as the source domain for model pretraining, and another domain $\mathcal{D}^{(t)}$ as the target domain for deployment. Under this protocol, performance is typically reported by averaging over a selected subset of such scenarios.

\section{Problem Formulation}

\begin{figure}[t]
    \centering
    \includegraphics[width=\linewidth]{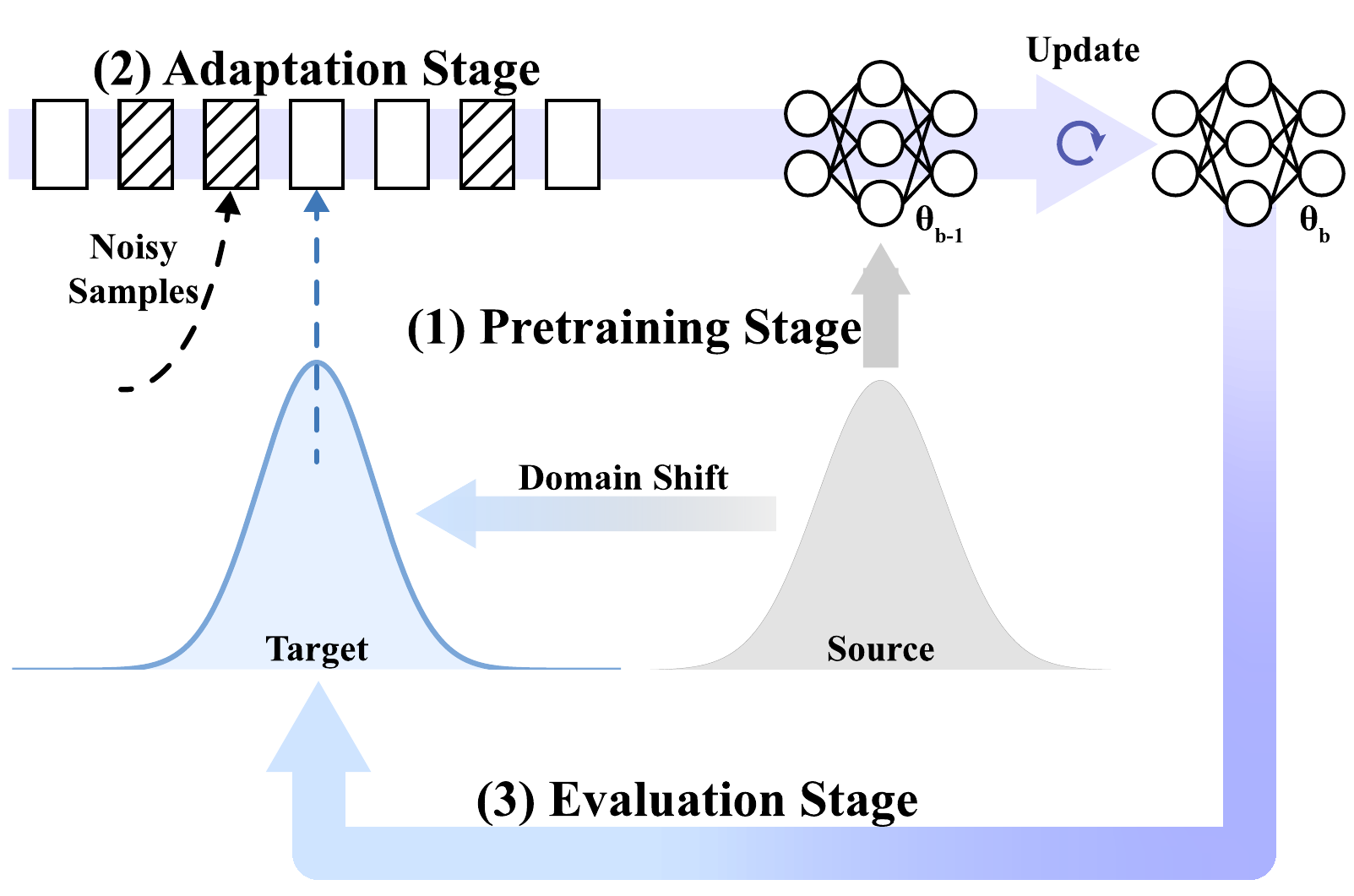}
    \caption{Illustration of the robust TTA setting. The model is pretrained on the source domain, adapted online on a mixed target stream containing both benign and noisy target samples, and finally evaluated only on the benign target domain.}
    \label{fig:setting}
\end{figure}

Standard TTA evaluates adaptation on a fixed source-target pair. To assess robustness under more realistic deployment conditions, robust TTA \cite{006_gong2023sotta, 007_lee2023towards, 008_gao2024unified, 009_park2024medbn} extends this setting by considering contaminated target streams. In one representative visual setting, a model pretrained on clean source data is adapted to a corrupted target benchmark, such as C-style datasets (e.g., CIFAR10-C, CIFAR100-C, and ImageNet-C \cite{011_hendrycks2019benchmarking}) with corruption shifts including weather-related corruptions like fog, snow, and frost. During test-time adaptation, additional noisy samples are mixed with benign target samples to construct a noisy online stream. The model is adapted on this mixed stream, while performance is evaluated only on the original benign target samples, as shown in Fig.~\ref{fig:setting} so as to assess whether the method can maintain target-domain performance despite noisy-stream interference. Following the contaminated-stream evaluation paradigm of SoTTA, we introduce stream contamination into the source–target scenarios commonly used in time-series TTA studies.

For each source–target scenario, we construct an online adaptation stream by combining benign target samples with one type of contamination. We consider five separate protocols. Attack contains adversarial samples designed to interfere with online adaptation. Near OOD and Far OOD introduce samples from external distributions that are, respectively, relatively close to and substantially different from the benign target distribution. Noise contains samples generated through predefined noise processes. The model is adapted on the resulting contaminated stream, whereas its performance is evaluated on benign target samples. Perturb. applies ten representative time-series perturbations to target
samples, with each perturbation evaluated at five severity levels. Dataset-specific OOD sources and the detailed generation procedures are provided in the experimental setup.

\section{Method}
\subsection{Motivation}
\begin{figure}[t]
    \centering

    \includegraphics[width=0.49\linewidth]{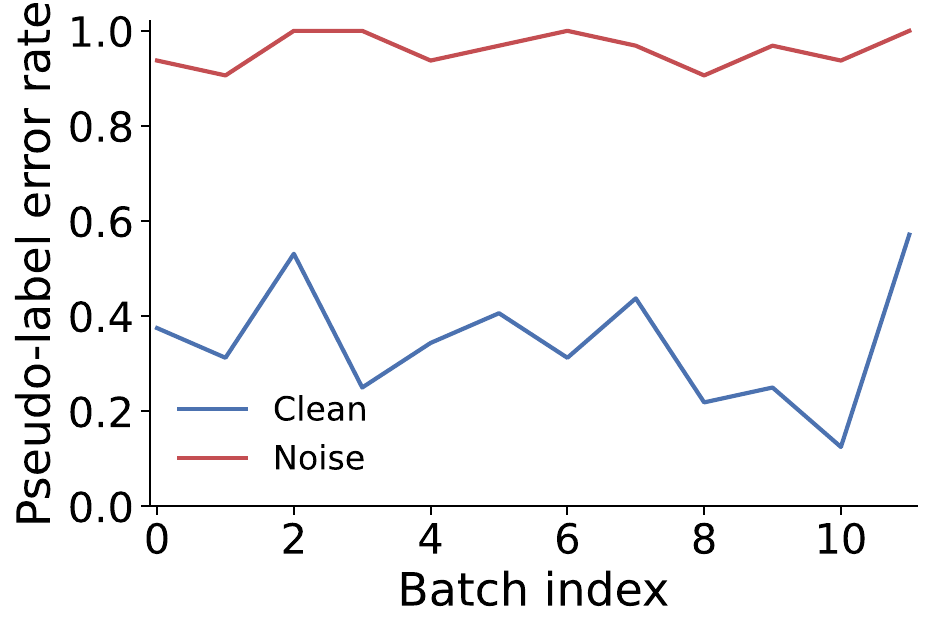}
    \hfill
    \includegraphics[width=0.49\linewidth]{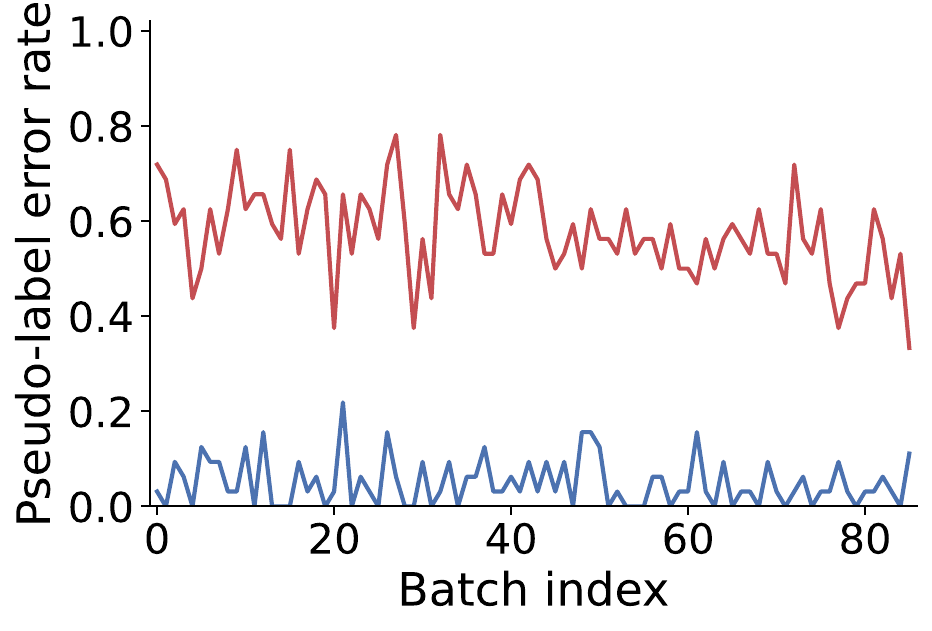}

    \caption{
    A representative case for pseudo-label error rates across streaming batches under clean and noisy conditions.
    (a) HAR.
    (b) FD.
    }
    \label{fig:pl_error}
\end{figure}

In streaming TTA, model updates are performed online using successive unlabeled batches. Consequently, unreliable supervision in the adaptation stream may directly affect model updates. Many existing TTA objectives rely on explicit pseudo labels to guide model adaptation. To illustrate the potential unreliability of such supervision under stream contamination, we examine pseudo-label quality in a representative case. As shown in Fig.~\ref{fig:pl_error}, the examined contaminated streams exhibit higher pseudo-label error rates than their benign counterparts. For pseudo-label-driven objectives, an incorrectly predicted class assignment is used as an explicit optimization target and may therefore provide a misleading adaptation signal. \textit{Motivated by this observation, we develop a pseudo-label-free objective that models sample relations without using predicted class assignments as explicit optimization targets.}

\begin{figure}[t]
    \centering
    \includegraphics[width=0.31\textwidth]{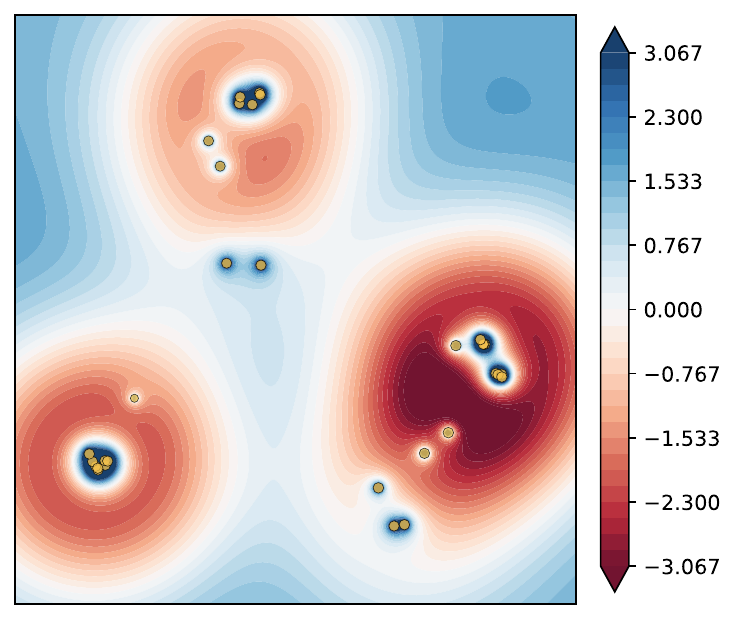}
    \hfill

    \includegraphics[width=0.32\textwidth]{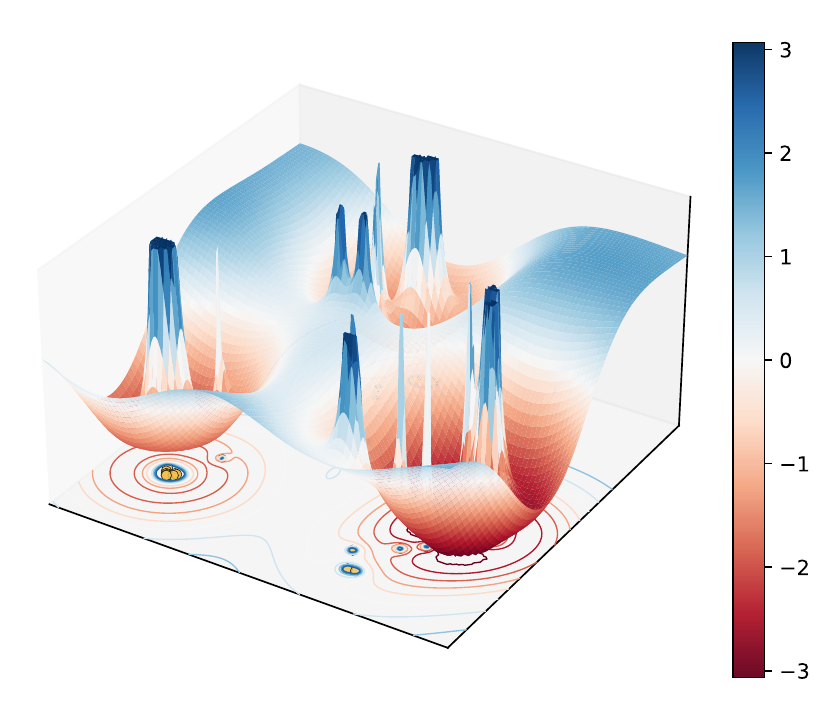}
    \caption{Visualization of the proposed potential field.
    (a) 2D contour map of the quality-aware potential field in the representation space.
    (b) 3D surface view of the same field.}
    \label{fig:potential_field}
\end{figure}

\subsection{Global Gravitational Potential Loss (GGPot)}

Inspired by the intuition of gravitational interaction, where attraction becomes stronger as object masses increase and their distance decreases, we formulate GGPot by modeling each target sample in the current batch as a particle in the representation space. In this view, the normalized feature vector determines the particle position, while the entropy-derived sample quality acts as its mass, so that more reliable samples exert stronger influence on the overall interaction field. GGPot combines a quality-aware attraction term with a mild repulsive regularizer to promote a compact yet non-collapsed representation structure.

\subsubsection{Batch-Wise Representation Construction}
For the current target batch $\mathcal{B}_b=\{X_{T,b}^j\}_{j=1}^{m_b}$, the model extracts its feature and prediction,
\begin{equation}
    \begin{aligned}
    z_b^j&=f_\psi(X_{T,b}^j)\in\mathbb R^d,
    \\
p_b^j&=h_\phi(z_b^j)=g_\theta(X_{T,b}^j)\in\mathbb R^C,
\end{aligned}
\end{equation}
where $z_b^j$ and $p_b^j$ denote the feature vector and logit vector, respectively. Here, $d$ represents the dimension of the feature vector, and $C$ represents the number of classes. Then, the feature vector is normalized to define the pairwise relation in the loss:
\begin{equation}
    \bar z_b^j=\dfrac{z_b^j}{||z_b^j||_2}.
\end{equation}

\subsubsection{Sample Quality Estimation}
To determine how important each sample is, we formulate an approach to estimate the quality $\omega_b^j$, corresponding to the mass in Newton's law of universal gravitation. A normalized entropy $\bar{H}_b^j$ is defined as,
\begin{equation}
    \begin{aligned}
        \pi_{b,c}^j&=\frac{\exp(p_{b,c}^j/\tau)}{\sum_{c'=1}^{C}\exp(p_{b,c'}^j/\tau)}
        \\
        H_b^j&=-\sum_{c=1}^{C}\pi_{b,c}^j\log(\pi_{b,c}^j+\epsilon),
\\
\bar H_b^j&=\frac{H_b^j}{\log C}.
    \end{aligned}
\end{equation}
We first convert the logit vector $p_b^j$ into a temperature-scaled predictive distribution. $\epsilon$ is a small positive constant added to avoid numerical issues. The softmax temperature $\tau$ controls the sharpness of the predicted probability distribution. Lower $\tau$ produces sharper, more peaked distributions, while higher $\tau$ yields smoother distributions. The final quality $\omega_b^j$ is defined as,
\begin{equation}
    \begin{aligned}
        q_b^j&=1-\bar{H}_b^j,\\
\omega_b^j&=\sigma\!\left(\kappa \left(q_b^j-\tfrac{1}{2}\right)\right),
    \end{aligned}
\end{equation}
where $\tfrac{1}{2}$ is the midpoint of the normalized range $[0,1]$, and $\kappa$ is a fixed slope parameter, set to $5$ in all experiments. Here, $q_b^j$ is defined as one minus the normalized entropy so that higher-confidence predictions correspond to higher quality scores, while uncertain predictions receive lower scores. As the temperature $\tau$ already affects the entropy distribution through softmax calibration, the subsequent sigmoid is used only for a fixed smooth calibration of the normalized quality score. Here, $0.5$ is the midpoint of the $[0, 1]$ range, and $5$ controls the steepness of the mapping. Both constants are fixed in all experiments.

\subsubsection{Formulation of GGPot}
Based on the reconstruction of the feature representation $\bar{z}_b^j$ and definition of the sample quality $\omega_b^j$, the distance between two samples $X_{T,b}^j$ and $X_{T,b}^k$, can be represented as
\begin{equation}
    d_b^{jk}=||\bar z_b^j-\bar z_b^k||_2.
\end{equation}
Accordingly, the total pairwise attraction among samples within a batch $b$ is defined as,
\begin{equation}
    \mathcal L_{\mathrm{att}}^{(b)}=-\frac{1}{2}
\sum_{j\neq k}
\omega_b^j\omega_b^k \exp\left(-\frac{(d_b^{jk})^2}{2\sigma^2}\right).
\end{equation}
The negative sign ensures that, under loss minimization, sample pairs with higher attraction are encouraged to move closer in the representation space. The summation over $j\neq k$ considers all distinct sample pairs, while the factor $\frac{1}{2}$ avoids double-counting symmetric pairs. The quality weights $\omega_b^j\omega_b^k$ make the attraction quality-aware, so that high-quality samples contribute more strongly to the attraction process. The Gaussian kernel assigns stronger attraction to nearby samples and progressively weaker attraction to distant ones, with $\sigma$ controlling the spatial range of this interaction. Since the attraction strength is modulated by both feature distance and sample quality, the objective primarily promotes local aggregation among reliable and already similar samples, rather than enforcing indiscriminate global contraction.

However, the attraction term alone may over-concentrate the representations and lead to a degenerate collapsed solution. To alleviate this issue, we introduce a small repulsive term
\begin{equation}
    \mathcal L_{\mathrm{rep}}^{(b)}=\frac{\lambda_{\mathrm{rep}}}{2}
\sum_{j\neq k}
\omega_b^j\omega_b^k
\frac{1}{d_b^{jk}+\epsilon},
\end{equation}
where $\lambda_{\mathrm{rep}}$ controls the strength of the repulsion. The inverse-distance form assigns strong penalties to excessively close sample pairs, while its effect decays as the distance increases, making it suitable for preventing collapse without disrupting the overall attraction structure. Here, $\epsilon$ is a small positive constant added to avoid numerical issues when the distance $d_b^{jk}$ between two samples is too small. Similar to the attraction term, the product $\omega_b^j\omega_b^k$ makes the interaction quality-aware, so that reliable samples contribute more strongly. The summation is taken over all distinct sample pairs, and the factor $\frac{1}{2}$ avoids double-counting symmetric pairs.

Finally, the total \textbf{GGPot} for batch $b$ is denoted as,
\begin{equation}
    \mathcal L^{(b)}=\frac{\mathcal L_{\mathrm{att}}^{(b)}+\mathcal L_{\mathrm{rep}}^{(b)}}
{m_b(m_b-1)},
\end{equation}
where $m_b$ denotes the number of samples in batch $b$. The normalization by $m_b(m_b-1)$ makes the objective less sensitive to batch size and allows it to be interpreted as the average pairwise interaction over distinct sample pairs.

From the potential-field perspective, Fig.~\ref{fig:potential_field} provides a visualization of the corresponding energy surface. It shows how quality-aware attraction creates favorable low-energy regions, whereas the repulsive component prevents the minimum from collapsing into an excessively concentrated configuration.

\paragraph{Graph-Energy Interpretation.}
Beyond the gravitational intuition, the pairwise formulation of GGPot also admits a natural graph-energy interpretation, where batch samples are viewed as nodes and their pairwise interactions define implicit weighted relations. In particular, for locally interacting pairs with $d_b^{jk}/\sigma$ sufficiently small, a Taylor expansion around $d_b^{jk}=0$ gives
\begin{equation}
\begin{aligned}
&-\omega_b^j\omega_b^k
\exp\left(-\frac{(d_b^{jk})^2}{2\sigma^2}\right) \\
&\quad =
-\omega_b^j\omega_b^k
+\frac{\omega_b^j\omega_b^k}{2\sigma^2}(d_b^{jk})^2
+\mathcal{O}\!\left(\frac{(d_b^{jk})^4}{\sigma^4}\right).
\end{aligned}
\end{equation}
The first term is independent of the pairwise distance, while the leading distance-dependent term takes the form of a weighted graph Dirichlet energy,
\begin{equation}
\sum_{j<k}
\omega_b^j\omega_b^k
\left\|\bar z_b^j-\bar z_b^k\right\|_2^2.
\end{equation}
Thus, the Gaussian attraction locally exhibits a quality-weighted graph-smoothing behavior.

\section{Experiments}

In this section, we evaluate GGPot within the ACCUP adaptation procedure and compare it with representative test-time adaptation methods on two time-series classification benchmarks.
Additional sensitivity analyses of the key hyperparameters and ablation experiments are provided in the Appendix.

\subsection{Experimental Setup}

\paragraph{Datasets}
We conduct the experiments on two commonly used real-world time series datasets: human activity recognition (UCIHAR) \cite{1_anguita2013public}, and machine fault diagnosis (MFD) \cite{26_lessmeier2016condition}. We select UCIHAR and MFD to cover two distinct real-world time-series applications, human activity recognition and machinery fault diagnosis, with different sensing modalities and domain-shift characteristics.

\begin{table*}[t]
\centering
\small

\begin{tabular*}{0.85\textwidth}{
@{\extracolsep{\fill}}
llccccc|cc
}
\toprule
& Method
& Attack
& Near
& Far
& Noise
& Perturb.
& Avg. $\uparrow$
& Worst $\uparrow$ \\
\midrule

\multirow{5}{*}{\textit{General TTA}}
& PL
& 81.29 & 81.53 & 81.34 & 81.34 & 81.43 & 81.39 & 81.29 \\

& SHOT
& 82.37 & \underline{82.45} & 82.49 & 82.42 & 82.42 & 82.43 & 82.37 \\

& T3A
& 82.26 & 78.08 & 84.45 & 84.02 & 80.12 & 81.79 & 78.08 \\

& TAST
& 81.99 & 78.56 & 84.06 & 83.84 & 78.23 & 81.34 & 78.23 \\

& TSD
& 80.76 & 82.05 & 80.96 & 80.94 & 81.47 & 81.24 & 80.76 \\

\midrule

\multirow{2}{*}{\textit{Robust TTA}}
& RoTTA
& 84.77 & 65.82 & \underline{84.91} & 84.62 & 49.52 & 73.93 & 49.52 \\

& SoTTA
& 82.42 & 82.42 & 82.42 & 82.42 & \underline{82.62} & 82.46 & \underline{82.42} \\

\midrule

\multirow{2}{*}{\textit{Time-Series TTA}}
& ACCUP
& \textbf{86.95} & 82.33 & 82.79
& \underline{86.03} & 81.17 & \underline{83.85} & 81.17 \\

& ACCUP \textbf{w/ Ours}
& \underline{86.51} & \textbf{82.66} & \textbf{85.24}
& \textbf{86.86} & \textbf{82.79} & \textbf{84.81} & \textbf{82.66} \\

\bottomrule
\end{tabular*}

\caption{Macro-F1 (\%) of TTA methods under five contaminated-stream
evaluation protocols for UCIHAR. Methods are grouped by their primary design setting:
general TTA, robust TTA, and time-series TTA. Avg. and Worst denote the
average and minimum performance across the five protocols, respectively.
Higher is better.}
\label{tab:main_ucihar}
\end{table*}

\begin{table*}[t]
\centering
\small

\begin{tabular*}{0.85\textwidth}{
@{\extracolsep{\fill}}
llccccc|cc
}
\toprule
& Method
& Attack
& Near
& Far
& Noise
& Perturb.
& Avg. $\uparrow$
& Worst $\uparrow$ \\
\midrule

\multirow{5}{*}{\textit{General TTA}}
& PL
& 81.14 & \underline{81.14} & 81.14 & 81.14 & 79.37 & \underline{80.79} & 79.37 \\

& SHOT
& 79.09 & 74.42 & 56.06 & 72.27 & 63.30 & 69.03 & 56.06 \\

& T3A
& 62.04 & 60.29 & 57.64 & 62.49 & 63.23 & 61.14 & 57.64 \\

& TAST
& 64.93 & 61.54 & 58.98 & 62.24 & 62.18 & 61.97 & 58.98 \\

& TSD
& 80.58 & 80.88 & 80.85 & 80.78 & 80.85 & \underline{80.79} & 80.58 \\

\midrule

\multirow{2}{*}{\textit{Robust TTA}}
& RoTTA
& \underline{91.09} & 37.45 & 61.99 & \underline{84.82} & 43.06 & 63.67 & 37.45 \\

& SoTTA
& 81.10 & 81.12 & 81.12 & 81.11 & \underline{81.15} & 81.12 & \underline{81.10} \\

\midrule

\multirow{2}{*}{\textit{Time-Series TTA}}
& ACCUP
& 84.60 & 77.28 & \underline{82.23}
& 76.48 & 80.78 & 80.27 & 76.48 \\

& ACCUP \textbf{w/ Ours}
& \textbf{91.97} & \textbf{93.61} & \textbf{93.27}
& \textbf{93.07} & \textbf{92.91} & \textbf{92.97} & \textbf{91.97} \\

\bottomrule
\end{tabular*}

\caption{Macro-F1 (\%) of TTA methods under five contaminated-stream
evaluation protocols for MFD. Methods are grouped by their primary design setting:
general TTA, robust TTA, and time-series TTA. Avg. and Worst denote the
average and minimum performance across the five protocols, respectively.
Higher is better.}
\label{tab:main_mfd}
\end{table*}

\paragraph{Noisy Datasets}
In addition to the benign target data introduced above, we consider five contaminated-stream protocols: RealWorld HAR~\cite{10_sztyler2016body}/CWRU~\cite{11_cwru_bearing_data} (Near), ESC-50~\cite{12_piczak2015esc} (Far), adversarial target samples (Attack), synthetic uniform-random time series (Noise), and target
samples affected by ten representative perturbations (Perturb.). Since RealWorld HAR and the target HAR benchmark both contain sensor-based human-activity time series, we use the running and jumping samples from RealWorld HAR as Near OOD data for UCIHAR. For MFD, we use the Ball-fault samples from CWRU as Near OOD data because both datasets contain vibration signals for rotating-machinery fault diagnosis. We use environmental-audio signals from ESC-50 as Far OOD data because their semantic content and physical generation process differ substantially from those of both human-activity and machinery-fault signals. Following SoTTA~\cite{006_gong2023sotta}, we use the terms Near and Far to distinguish external samples with different degrees of relatedness to the benign target data. For Attack, we adopt the Distribution Invading Attack (DIA), which is an adversarial attack algorithm designed for TTA~\cite{_40_wu2023uncovering}. We optimize small perturbations on the malicious samples to disrupt the adaptation process and increase the classification loss on the benign samples in the same batch. For Noise, we independently generate element-wise uniform-random time series with the same number of channels and temporal length as the target inputs. In contrast, Perturb. applies ten representative time-series perturbations at five severity levels to the original target samples and mixes the resulting perturbed samples into the adaptation stream. By default, each adaptation batch contains equal numbers of benign and contaminated samples, corresponding to a benign-to-contamination ratio of 1:1. All contaminated samples are transformed to the target input shape and processed using the same preprocessing pipeline as the corresponding benign target data, without contamination-specific normalization.

\paragraph{Baselines}
We implement GGPot within the ACCUP adaptation procedure and compare it with the original ACCUP~\cite{005_gong2025augmented} and other
representative general TTA baselines, including PL~\cite{25_lee2013pseudo}, SHOT~\cite{30_liang2020we},
T3A~\cite{17_iwasawa2021test}, TAST~\cite{18_jang2022test},
TSD~\cite{56_wang2023feature}. We also compare with the robust
TTA methods RoTTA~\cite{65_yuan2023robust} and
SoTTA~\cite{006_gong2023sotta}.

\subsection{Comparative Experiments}
The proposed method is evaluated on UCIHAR and MFD. Tables~\ref{tab:main_ucihar} and \ref{tab:main_mfd} report the results on UCIHAR and MFD, respectively, under the five contaminated-stream protocols, together with the average and worst-case performance. Each score is averaged over five scenarios run with three seeds.

\paragraph{UCIHAR Dataset}
Table~\ref{tab:main_ucihar} reports the Macro-F1 scores on UCIHAR under the five contaminated-stream protocols. ACCUP w/ Ours achieves the best average performance of 84.81\%, outperforming the original ACCUP by 0.96 percentage points. It improves ACCUP under the Near, Far, Noise, and Perturb. protocols, with the largest gain of 2.45 percentage points under Far OOD. Although its performance under Attack decreases slightly from 86.95\% to 86.51\%, it remains competitive and achieves the best results under the other four protocols. Moreover, ACCUP w/ Ours improves the worst-case performance from 81.17\% to 82.66\%, which is also the highest among all compared methods. These results demonstrate that the proposed strategy improves the overall and worst-case performance of ACCUP across the five contaminated-stream protocols.

\paragraph{MFD Dataset}
Table~\ref{tab:main_mfd} reports the Macro-F1 scores on MFD under the five contaminated-stream protocols. ACCUP w/ Ours achieves the best average performance of 92.97\%, outperforming the original ACCUP by 12.70 percentage points. It ranks first under all five protocols, with improvements over ACCUP ranging from 7.37 percentage points under Attack to 16.59 percentage points under Noise. Moreover, the proposed strategy substantially improves the worst-case performance of ACCUP from 76.48\% to 91.97\%, which is also the highest among all compared methods. These results demonstrate substantial and consistent improvements across heterogeneous contaminated streams, in terms of both average and worst-case performance.

\subsection{Semantic Enrichment of Attractive Interactions} \label{sec:homophily}
The above results demonstrate the effectiveness of GGPot under
contaminated adaptation streams. To better understand these results, we examine whether GGPot assigns more attractive weight to same-class pairs.
\paragraph{Assumption 1.}
We assume that the attractive interactions of GGPot favor same-class
pairs beyond what is expected from random pairing. Let $\pi_c$ denote the proportion of class $c$. We define the class-prior baseline as
\begin{equation}
H_{\mathrm{prior}}=\sum_{c=1}^{C}\pi_c^2,
\end{equation}
which gives the expected same-class probability under random pairing. To characterize the induced pairwise relations, we further define
\begin{equation}
\begin{aligned}
H_{\mathrm{distance}}
&=
\frac{
\sum_{j<k} K_{jk}\mathbb{I}[y_j=y_k]
}{
\sum_{j<k} K_{jk}
},\\
H_{\mathrm{weighted}}
&=
\frac{
\sum_{j<k} \omega_j\omega_k K_{jk}\mathbb{I}[y_j=y_k]
}{
\sum_{j<k} \omega_j\omega_k K_{jk}
}.
\end{aligned}
\end{equation}
where
$K_{jk}=\exp\left(-\|\bar z_j-\bar z_k\|_2^2/(2\sigma^2)\right)$
denotes the Gaussian distance weight. Assumption~1 therefore corresponds to $H_{\mathrm{weighted}}>H_{\mathrm{prior}}$.

\begin{table}[t]
\centering
\small
\setlength{\tabcolsep}{4.5pt}
\begin{tabular}{@{}llccc@{}}
\toprule
Dataset & Setting & $H_{\mathrm{prior}}$
& $H_{\mathrm{distance}}$
& $H_{\mathrm{weighted}}$ \\
\midrule

\multirow{5}{*}{MFD}
& Attack & \multirow{5}{*}{0.421} & 0.761 & 0.782 \\
& Clean  &                         & 0.864 & 0.898 \\
& Far    &                         & 0.775 & 0.782 \\
& Near   &                         & 0.759 & 0.785 \\
& Noise  &                         & 0.822 & 0.839 \\
\midrule

\multirow{5}{*}{UCIHAR}
& Attack & \multirow{5}{*}{0.170} & 0.854 & 0.885 \\
& Clean  &                         & 0.899 & 0.928 \\
& Far    &                         & 0.785 & 0.801 \\
& Near   &                         & 0.649 & 0.673 \\
& Noise  &                         & 0.894 & 0.923 \\
\bottomrule
\end{tabular}
\caption{Semantic enrichment of attractive interactions across different settings. Higher values indicate a larger fraction of interaction weight assigned to same-class pairs.}
\label{tab:semantic_enrichment}
\end{table}

\paragraph{Results}
Table~\ref{tab:semantic_enrichment} reports the homophily results under the clean-stream setting and four representative contaminated-stream protocols: Attack, Near, Far, and Noise.
For both datasets, $H_{\mathrm{distance}}$ consistently exceeds
$H_{\mathrm{prior}}$ across all reported settings, indicating that nearby
representations are already enriched for same-class relations.
More importantly, $H_{\mathrm{weighted}}$ remains above
$H_{\mathrm{prior}}$ in every setting, supporting Assumption~1.
Moreover, $H_{\mathrm{weighted}}>H_{\mathrm{distance}}$ throughout
the reported results, suggesting that quality weighting
further strengthens the enrichment of same-class interactions.

\begin{figure}[t]
    \centering
    \includegraphics[width=\linewidth]{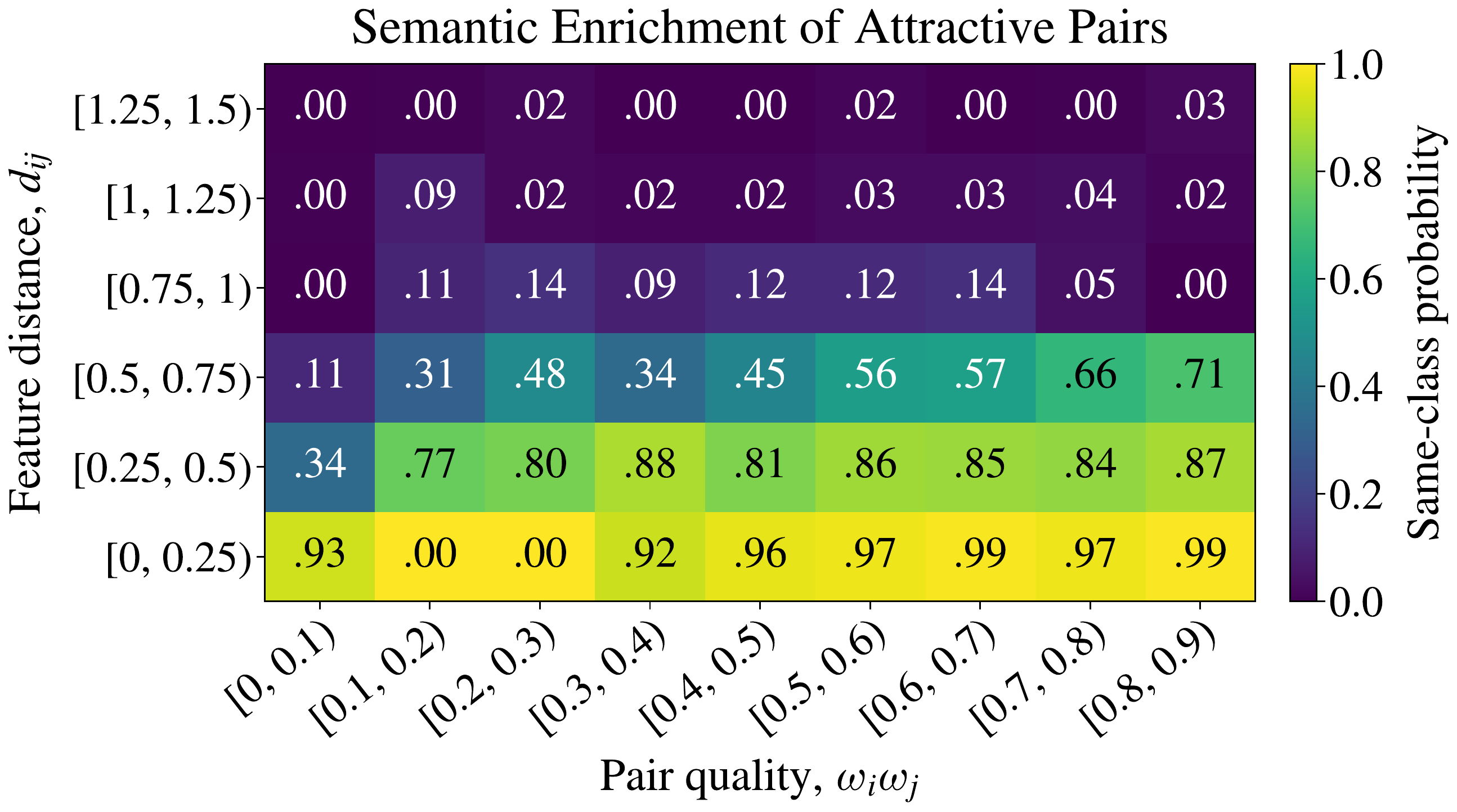}
    \caption{Fine-grained semantic enrichment of attractive interactions. Each cell shows the attraction-mass-weighted same-class probability within a pair-quality--distance bin.}
    \label{fig:semantic_enrichment_heatmap}
\end{figure}

Figure~\ref{fig:semantic_enrichment_heatmap} shows that same-class probability is generally higher for pairs with larger quality weights and smaller feature distances. This suggests that distance-based attraction favors semantically related local pairs, while quality weighting further emphasizes reliable interactions.

\section{Conclusion}
In this paper, we revisited time-series test-time adaptation from a robustness-oriented perspective. We introduced five contaminated-stream evaluation protocols to assess robustness under various adaptation interference, and complemented standard cross-domain evaluation with contaminated-stream protocols that assess the stability of online adaptation under heterogeneous interference. To reduce reliance on potentially erroneous class-valued supervision, we proposed GGPot, a pseudo-label-free, quality-weighted objective. Experiments on UCIHAR and MFD showed that ACCUP with GGPot improves the average and worst-case performance over the original ACCUP on both datasets and achieves the highest average performance among the evaluated methods.

\section{Generative AI Use Disclosure}
Generative AI tools assisted with language editing. All content was verified by the authors.

\bibliography{sample-base}

\end{document}